%% file: main.tex
\documentclass[11pt,3p,review]{elsarticle}

\makeatletter
\def\ps@pprintTitle{%
 \let\@oddhead\@empty
 \let\@evenhead\@empty
 \def\@oddfoot{}%
 \let\@evenfoot\@oddfoot}
\makeatother

\usepackage{xurl}
\usepackage{adjustbox}
\usepackage{multirow}
\usepackage{xcolor}
\usepackage{lineno}
  
\begin{document}
    
\begin{frontmatter}

\title{Streamlined Framework for Agile Forecasting Model Development towards Efficient Inventory Management}

\author[aic]{Jonathan~Hans~Soeseno}
\author[aic]{Sergio~Gonz\'alez}
\author[aic]{Trista~Pei-Chun~Chen}


\address[aic]{AI Center, Inventec Corporation, Taipei, Taiwan}





\begin{abstract}
This paper proposes a framework for developing forecasting models by streamlining the connections between core components of the developmental process. The proposed framework enables swift and robust integration of new datasets, experimentation on different algorithms, and selection of the best models. We start with the datasets of different issues and apply pre-processing steps to clean and engineer meaningful representations of time-series data. To identify robust training configurations, we introduce a novel mechanism of multiple cross-validation strategies. We apply different evaluation metrics to find the best-suited models for varying applications. One of the referent applications is our participation in the intelligent forecasting competition held by the United States Agency of International Development (USAID). Finally, we leverage the flexibility of the framework by applying different evaluation metrics to assess the performance of the models in inventory management settings.

\end{abstract}



\end{frontmatter}


\input{sections/01_introduction}
\input{sections/02_related_work}
\input{sections/03_method}
\input{sections/04_case_study}
\input{sections/05_experiments}

\input{sections/06_conclusion}


\bibliography{refs}

\end{document}

%% file: sections/01_introduction.tex
\section{Introduction}

Logistics is defined as the coordination of information, physical, and financial flows to and from trading partners and consumers. An inventory requires capital investment to build and stock finished goods. Inventory management considers several complex variables, including time, cost, location, transport, and risks. Logistics inventory management is a long-standing challenge that is essential for various aspects of life. One concrete example is its impact on global health \citep{bhaskar2020epicenter,rostami2021forecasting}. In this setting, adequate inventory management translates to demand satisfaction of medical products, which could potentially save lives, prevent unwanted pregnancies, and improve the general well-being of people's lives. Another example in an industrial setting can be drawn by considering a manufacturing company that further assists companies in manufacturing their products \citep{syntetos2016supply}. Here, each company may request hundreds or even thousands of products. Inventory management plays a crucial role in reducing costs and mitigating the loss of potential income. Many factors affect the results of inventory management solutions. One prominent factor is the ability to forecast future demand accurately. Accurate demand forecasts, in turn, facilitate better management of inventory logistics.

Therefore, forecasting algorithms may be used to perform the analysis. As a considerably popular field, there are many state-of-the-art algorithms that are mainly categorized into statistical and AI-based approaches. Statistical approaches often focus on modeling the temporal relation of a given time-series data \citep{box2015time,HA14}. Thus, they are robust and accurate. However, these approaches rely heavily on data pre-processing steps to ensure the underlying statistical constraints, such as stationarity (i.e., time-series statistics are constant over time), seasonality, and the trend of an autoregressive integrated moving average (ARIMA) model \citep{box2015time}. Because of this requirement, statistical approaches are often used to model specific time-series data and are therefore difficult to maintain at scale.

By contrast, AI-based approaches utilize machine learning and deep learning models to solve tasks that may involve larger datasets \citep{jordan2015machine,goodfellow2016deep,lim2021time}. Recently, AI-based forecasting approaches have become more prominent in accommodating multiple time series using a single model \citep{bojer2021kaggle,makridakis2021m5B}. This technique, known as cross-learning, allows the model to share helpful information across time-series data \citep{semenoglou2021investigating}. It also simplifies the development and deployment processes. However, similar to all other deep learning approaches, AI-based forecasting approaches struggle with smaller datasets, creating the necessity to tune and hand-engineer complex data representations.

With such a wide variety of algorithm choices available, the problem of maximizing performance then translates to searching for the most suitable algorithm for the task at hand, often described with a recorded dataset. Unfortunately, developing a forecasting model is not trivial, mainly because of the complexity of its components. For instance, the datasets may consist of various issues and formats, while the forecasting algorithms require specific feature representations and proper hyperparameter tuning \citep{cawley2010over,probst2019tunability}. In addition, it would still be necessary to define a metric to select the best model \citep{cerqueira2020evaluating,koutsandreas2021selection}, which varies significantly depending upon the intended application of the models. For example, custom evaluation metrics based on an inventory simulation process may be considered to estimate the performance of an accurate model given an inventory management setting.

To address these issues, we propose a streamlined framework that organizes each component in a modular manner. Therefore, it allows users to quickly iterate through combinations of algorithms and datasets in a one-to-many or many-to-many fashion. With numerous possible combinations of feature representations, model architecture, training setup, and other relevant hyperparameters, a robust strategy is required to choose models with the best generalization properties. Hence, the framework enables the use of novel multiple cross-validation strategies built on top of existing ones. In addition, given the flexibility of the framework in the application of various evaluation metrics with guaranteed reliability and reproducibility, it enables us to extract insights from each model and select the most suitable model or improve the model design depending on the requirements. 


To evaluate the effectiveness of our framework, we perform several experiments with four real-life forecasting problems. One of the referents is our participation in the intelligent forecasting competition organized by USAID. We begin by comparing the strategy of learning one model for each time series and cross-learning a single model to predict multiple series. Further, we empirically demonstrate the advantages of combining multiple cross-validation strategies to tune robust models with better generalization. Finally, we investigate the performance of such accurate forecasting models in the inventory management setting by applying metrics derived from an inventory simulation process. We summarize our contributions as follows:

\begin{itemize}
    \item A streamlined framework that allows users to quickly iterate through different AI model training configurations by swiftly integrating various datasets, algorithms, and metrics.
    \item A scheme to facilitate a robust search for the most optimal training configurations through the integration of multiple cross-validation strategies.
    \item A comprehensive evaluation of forecasting models based on user-defined metrics, including conventional and inventory specific metrics.
\end{itemize}

The remainder of this paper is organized as follows. Section \ref{sec:works} presents the background and related works on forecasting approaches, tuning and model selection, and inventory management techniques. Section \ref{sec:method} describes in detail the modules of the proposed streamlined forecasting framework, as well as its main strengths and applications. Section \ref{sec:case-study} discusses our study case on the intelligent forecasting competition by USAID. In Section \ref{sec:experiments}, we empirically analyze the design of our framework in terms of the use of cross-learning, integration of multiple cross-validation strategies, and its applicability to inventory management settings. Finally, Section \ref{sec:conclusion} concludes our study and discusses practical future work.

%% file: sections/02_related_work.tex
\section{Related works}
\label{sec:works}

Forecasting \citep{HA14,box2015time} is one of the most attractive research fields in statistics and machine learning because of its impact on organizational decision-making. In particular, forecasting is beneficial in financial, economic, and operational processes \citep{makridakis2020forecasting}, such as demand forecasting and inventory management \citep{syntetos2009forecasting,syntetos2016supply}. A vast majority of forecasting research is focused on the pursuit of economic profit. However, the knowledge learned in the past few years is also applicable to general and social goods \citep{rostami2021forecasting}. As forecasting practitioners, we have the responsibility of executing the same. For example, demand forecasting insights that are generally utilized by retailers or manufacturing companies were instead used for humanitarian aid in \citep{van2016demand, behl2019humanitarian, burba21}.

Therefore, practitioners aim to develop robust and accurate forecasting algorithms. Section \ref{sec:stat-ai-methods} addresses these forecasting approaches from the perspective of statistical and AI-based models. In Section \ref{sec:parameter-tuning}, we discuss the importance of hyperparameter tuning on model performance, as well as the metrics and validation strategies used in forecasting scenarios. Finally, Section \ref{sec:forecast-inventory} examines the proposals for the transition of forecasting models into inventory management decisions.

\subsection{Forecasting algorithms}
\label{sec:stat-ai-methods}

Over the years, the forecasting community has questioned which approaches are more suitable for different time series forecasting problems \citep{hong2019global,hyndman2020brief,makridakis2021m5B}. The superiority of statistical forecasting algorithms over AI-based approaches is still an open discussion \citep{gilliland2020value,spiliotis2020comparison}. The former includes classical and well-known approaches, such as ARIMA \citep{box2015time}, exponential smoothing \citep{gardner1985exponential,gardner2006exponential}, and Croston's methods \citep{croston1972forecasting,babai2019new}, among others. The latter group involves the use of deep learning \citep{goodfellow2016deep,lim2021time} and machine learning \citep{jordan2015machine} algorithms along with feature engineering \citep{guyon2008feature}. 

In this discussion, \citet{makridakis2018statistical} and \citet{gilliland2020value} based on the results of the M4 competition, claimed that the more complex AI-based approaches do not necessarily outperform simpler statistical forecasting models. However, more recent studies \citep{spiliotis2020comparison,bojer2021kaggle} and the M5 competition \citep{makridakis2021m5A,makridakis2021m5B} demonstrated a superior performance by the AI-based models. Recent AI-based approaches have the ability to learn a single model from multiple time series data to forecast individual time series \citep{semenoglou2021investigating}. This mechanism, referred to as cross-learning, is one of the contributors to the improvement in performance in the latest competitions \citep{smyl2020hybrid,semenoglou2021investigating,bojer2021kaggle}. 

Among the AI-based approaches, recurrent neural networks \citep{lim2021time}, particularly long short-term memory (LSTM) \citep{hochreiter1997long}, and gradient boosting decision trees (GBDT) \citep{friedman01,friedman02}, in this case LightGBM \citep{ke2017lightgbm}, have been the most successful algorithms for forecasting problems \citep{bojer2021kaggle}. LSTMs are neural networks with added cells that facilitate learning from sequence or temporal data. LightGBM is Microsoft's GBDT implementation with leaf-wise tree growth and histogram-based cut selection, which makes it one of the most efficient and effective GBDTs \citep{gonzalez2020practical}.

In addition to statistical and AI-based models, hybrid algorithms combine classic forecasting approaches with the latest AI-based techniques \citep{smyl2020hybrid,montero2020fforma,triebe2020}. As a reference, NeuralProphet \citep{triebe2020,NP-soft} integrates the classical decomposition time series analysis of Facebook's Prophet \citep{taylor2018forecasting} and AR-Net \citep{triebe2019ar}. AR-Net is a neural network architecture that performs an autoregressive procedure similar to ARIMA but offers more options for scalability.

\subsection{Hyperparameter tuning: metrics and validation strategies}
\label{sec:parameter-tuning}

Forecasting algorithms, especially AI-based algorithms, such as LSTM or GBDT, consist of a considerable number of hyperparameters that are required to be tuned to access their full potential \citep{probst2019tunability}. Additionally, the correct selection of the trained models through different validation strategies is essential to achieve generalization and avoid overfitting \citep{cawley2010over}. This process can be considerably costly, both computationally and for practitioners who analyze the resultant models. LSTM and GBDT were not ranked as the best forecasting approaches until they were computationally optimized \citep{bojer2021kaggle}. 

Tunability and model selection is a complex stage of AI modeling with a considerable research interest, which has motivated meta-learning and AutoML \citep{hutter2019automated,zoller2021benchmark}. This interest is extendable to forecasting scenarios \citep{petropoulos2018judgmental,bakhashwain2021online}. However, it becomes even more difficult to forecast problems due to the lack of a clear consensus on the metrics and validation strategies. Additionally, the selection of engineered features commonly forms part of the tuning stage of AI forecasting models.   

Over the years, multiple metrics have been proposed to evaluate the accuracy of these forecasting models \citep{hyndman2006another,koutsandreas2021selection}, including scale-dependent regression metrics, such as mean absolute error (MAE) and mean squared error (MSE); percentage metrics, such as symmetric mean absolute percentage error (SMAPE); and, native forecasting metrics, such as mean absolute scaled error (MASE). In forecasting model validation, three different groups of strategies outshine the rest \citep{cerqueira2020evaluating}. Out-of-sample validation retains a fragment at the end of the time series as the validation set \citep{tashman2000out}. Prequential or time series cross-validation uses an increasing window, which defines the training and validation sets, and repeats the process several times while increasing the window sizes. K-fold cross-validation divides the time series into equal parts from which one is used for validation, while the rest is included in the training set \citep{bergmeir2012use,bergmeir2018note}.

\subsection{From forecasting to inventory management}
\label{sec:forecast-inventory}

Once the models are developed, the forecasts can be applied to inventory management decisions. This transition might not be straightforward, as the most accurate models do not necessarily lead to the most suitable inventory management \citep{kourentzes2020optimising,goltsos2021inventory,kourentzes2021connecting}. Therefore, various approaches have been proposed following two main paradigms, depending on whether they consider forecasting and inventory management separate problems or a single learning process \citep{huber2019data,goltsos2021inventory}. In the first paradigm, the demand distribution is generally assumed or roughly estimated using the forecasts and their errors \citep{silver2016inventory,huber2019data}. The second paradigm simultaneously optimizes the forecasting models and decisions. Some examples of these approaches are the minimization of the holding and shortage costs in newsvendor problems \citep{huber2019data,punia2020predictive}, consideration of demand uncertainty in Bayesian forecasting models \citep{prak2019general}, and inclusion of simulation optimization \citep{kourentzes2020optimising}. These approaches might not be as generalist as desired. That is, they usually are defined on a particular inventory setting, or limited to specific forecasting models.

Our approach is distinct from the existing inventory optimization approaches, as we do not impose any distributions like the first paradigm; nor do we perform any type of inventory optimization. Instead, we aim to select the best configurations of the forecasting models based on their performance in terms of inventory management metrics obtained through a simulation process.


%% file: sections/03_method.tex
\section{Method}
\label{sec:method}
We aim to enable an agile development of forecasting models on a large scale. However, it is a challenging task because the development process may require several stages to work coherently. Therefore, we have designed a streamlined framework that allows us to robustly and quickly develop our model by seamlessly integrating various datasets, model architectures, and evaluation metrics. 

We begin by designing pre-processing modules for each dataset that apply the most appropriate data cleaning procedures and enable a uniform interface across the datasets. Next, we engineer lagged, pattern, and statistical features to allow improved time-series data representation for the models and store the cleaned data alongside the engineered feature representations in our dataset repository. We further utilize the repository of datasets and algorithms to train our model and tune the hyperparameters to satisfy user-defined evaluation metrics. Figure \ref{fig:schematic_diagram} provides an overview of the developmental procedures of our framework.

The pre-processing step to enable a uniform and clean interface across different datasets is explained in Section \ref{sec:method_preprocessing}. Section \ref{sec:method_feature_engineering} addresses the utilization of this uniform interface to build improved feature sets to represent the time series. Next, Sections \ref{sec:method_training_tuning} and \ref{sec:method_evaluation_metric} discuss the procedure used to train and tune the models to satisfy the evaluation metrics. Finally, in Section \ref{sec:method_streamlined_framework}, we highlight the main strengths of the proposed streamlined framework and introduce real-world applications.

\input{figures/schematic_diagram}

\subsection{Uniform interface for datasets}
\label{sec:method_preprocessing}
In this step, the goal is to pre-process datasets obtained from various data sources, which are likely to consist of different sizes, formats, and issues. It is possible to design a single pipeline that is suitable for a majority of these datasets. However, doing so would introduce substantial complexity for accommodating large dataset variations. Such a tremendous effort would still not be useful when we consider new datasets. Therefore, to avoid additional and unnecessary complexity, we instead design a pre-processing module for each dataset.

These pre-processing modules, further referred to as data loaders, enable us to apply the most appropriate pre-processing steps for each dataset. For instance, removing missing values from one dataset, or designing imputation strategies on another dataset. However, because the datasets might contain different information presented in various formats, the subsequent modules would require a specific interface to access each of them. Therefore, in addition to applying cleaning and pre-processing steps, we design the data loaders to provide a uniform interface to access the time series identifier, timestep, and target. We consider these features essential while developing forecasting models.

\subsection{Descriptive representation of time series}
\label{sec:method_feature_engineering}

The uniform interfaces provided by the data loaders enable us to perform feature engineering over each dataset. This procedure aims to engineer meaningful representations that allow AI models to understand time-series data better. For instance, instead of asking the model to memorize the past eight recorded values, we can explicitly feed them as inputs. Doing so allows the AI model to focus more on interpreting, rather than memorizing. This strategy is beneficial while dealing with smaller datasets, as specific patterns might require larger datasets to extract. 



The feature engineering procedure considers three different types of features. The \textit{lagged features} are implemented by directly shifting the target variable backward in time. The \textit{pattern features} focus on describing the shape of the time-series data within a given window. Depending on the recording frequency of the dataset, the module extracts information, such as the percentage of zero values, last-observed non-zero values, trends, and occurrences of up-down patterns. Finally, we compute the \textit{statistic features} of the recorded values within a time series window, such as the mean, minimum, maximum, and standard deviation.


This step results in feature sets that are stored inside the dataset repository to accompany the cleaned dataset. It is possible to include other features obtained from external sources into this repository. One of such occasions was during our participation in the USAID intelligent forecasting competition, where we incorporated world population data in addition to the three time-series features.

\subsection{Training and tuning AI models}
\label{sec:method_training_tuning}
Given multiple datasets and state-of-the-art forecasting algorithms available in our repositories, we aim to develop a model that offers the best performance. To this end, we need to maximize the model's performance by tuning the hyperparameters that control its capacity, training strategy, learning objective, and overall performance. In addition to the large number of hyperparameters, we are also required to carefully select the features from the datasets, as some features may be crucial in improving the performance of the model, while others might introduce unnecessary intricacies.

Combining the variation of the hyperparameters with the dataset features could quickly generate a massive number of configurations. For example, consider ten hyperparameters with three possible values each. When we combine them with 40 unique feature sets, the number of combinations could easily reach millions. Given more complex model architectures with more hyperparameters, it would be impractical to exhaustively attempt all the configurations because of a significant computational resource and time requirement. Although impractical, it is still beneficial to optimize the training process and leverage from parallelization opportunities to sample and iterate through many configurations, albeit not all of them. By analyzing different configurations at such a scale, we can gain more insight into the problem and thus design better data representations or model architectures. 

\input{figures/cross_validation_illustration}

Therefore, it is necessary to select configurations that yield desirable performance and robust models. To do so, we incorporated two different validation strategies: time series and k-fold cross-validation (see the illustration in Figure \ref{fig:cross_validation_illustration}). The time series cross-validation is performed on the time axis. Our k-fold cross-validation is performed across different time series, contrary to traditional k-fold strategies. In each fold, we exclude certain time series from training to constitute the validation set.

The main idea of cross-validation is to evaluate the performance of a particular model against all seen portions of the datasets, leading to a more reliable performance estimation. However, using only one cross-validation strategy might not be sufficient when multiple time series exist. By exclusively using the time series cross-validation, we only evaluate the performance of the model in dealing with future unseen data points and ignore generalization over different patterns of other time series. In contrast, using only the k-fold cross-validation, we only examine the generalization on other time series. Therefore, to ensure the selection of robust models, we consider the application of both cross-validation strategies, simultaneously. We accommodate the top configurations that are highly ranked by time series and k-fold cross-validations. Further, we demonstrate the effectiveness of using multiple cross-validation strategies in Section \ref{sec:exp_validation_strategy}.

\subsection{Evaluation metric}
\label{sec:method_evaluation_metric}
Each training configuration yields a forecasting model. With a large pool of models obtained from the hyperparameter tuning process, we intend to reliably evaluate and select the best-performing model. To this end, we separate the evaluation metrics from the hyperparameter tuning process. This separation enables us to apply multiple evaluation metrics to a specific training configuration, and therefore, allows us to select the model that best fits the intended application. We include a few conventional metrics \citep{hyndman2006another,koutsandreas2021selection}, such as MAE, RMSE, and MASE, defined as follows:
\begin{equation}
    MAE = \frac{1}{N}\sum_{t=1}^{N}|y_t - \hat{y}_t|
\end{equation}

\begin{equation}
    RMSE = \sqrt{\frac{1}{N}\sum_{t=1}^{N}(y_t - \hat{y}_t)^2}
\end{equation}

\begin{equation}
    MASE = \frac{\frac{1}{N} \sum_{t=1}^{N}|y_t - \hat{y}_t|}{\frac{1}{N-1} \sum_{t=2}^{N}|y_t - y_{t-1}|}
\end{equation}

Here, $y_t$ and $\hat{y}_t$ denote the ground-truth target and model prediction at time $t$, respectively. These conventional metrics measure the accuracy of the models based on how close the predictions are to the ground-truth targets. In addition, the framework can accommodate more complex metrics on inventory performance \citep{cannon2008inventory}, such as the days-of-inventory (DOI) and stockout rate (SR). The DOI evaluates the average time that the inventory is stored before sale. The SR measures the ability to satisfy demand based on the available inventory. These are defined as follows: 
\begin{equation}
    \label{eq:doi}
    DOI = \frac{\mathit{Avg.\ Inventory\ Cost}}{\mathit{Cost\ of\ Goods\ Sold}} \times \mathit{Days}
\end{equation}
\begin{equation}
    \label{eq:sr}
    SR = \frac{\mathit{Unfulfilled\ Demand}}{\mathit{Total\ Demand}}
\end{equation}

We formulate an inventory simulation to compute these inventory metrics. The simulation considers the recorded sales at each time step as the optimal inventory replenishment. Under this condition, the inventory would be consumed and replenished with the same amount, leaving zero inventory at the end of every timestep. Thus, the replenishment is instant, similar to \citep{guo2014prediction}. From here, we can substitute the replenishment with the predicted sales at timestep while retaining the original sales as the consumption of the inventory. In this setting, we are able to obtain a more refined insight into how a specific model performs in a given inventory task. For example, underestimating future demands would lead to a higher SR, while overestimation would result in a higher DOI.

We specifically choose the DOI and SR metrics, as they represent a crucial trade-off in the inventory management setting. Note that minimizing either of the metrics would lead to a catastrophic impact on the other. For instance, while minimizing the DOI of the inventory level, there exists a degenerate solution with zero inventory level at all times. However, doing so would result in a very high SR. The opposite occurs when we attempt to minimize the SR. We can maintain the inventory at its maximum capacity to meet every demand, but doing so would result in an extremely high DOI. Figure \ref{fig:doi_stockout_tradeoff} provides a visual illustration of the DOI and SR trade-off, where Figure \ref{fig:doi_stockout_tradeoff} (a) represents the ideal trade-off, while Figure \ref{fig:doi_stockout_tradeoff} (b) and (c) represent the trivial cases.  

\input{figures/doi_stockout_tradeoff}

\subsection{Streamlined AI model framework}
\label{sec:method_streamlined_framework}

Given the model development components, we arrange them into a streamlined framework, beginning from pre-processing raw datasets, up until the evaluation metrics. We further optimize the framework to quickly iterate through multiple training configurations involving various model hyperparameters and feature sets. This functionality enables three different application scenarios.

The first scenario compares several models over a specific dataset to search for the model with the highest prediction accuracy. One of the actual applications of this scenario is our participation in the USAID intelligent forecasting competition, where we were required to fit and compare as many models as possible over a single dataset. In Section \ref{sec:experiments}, we discuss the design choices for our best model. Another scenario is when a new model architecture is included in the model repository. Before adding such a model to our collection, we aim to understand its behavior in different settings of the datasets. In this case, we would obtain a single model fitted over several datasets. The last scenario is to benchmark different models over different datasets in an N-model and N-dataset fashion. This scenario is primarily applicable when writing research papers in which such a comprehensive benchmark is important to highlight the contribution of the study.

These three different applications of the streamlined framework contribute to its versatility and applicability in multiple occasions. This versatility eventually led to the development of new models, which were then added back into the repository, further enriching our model collection. Therefore, prototyping solutions for future projects becomes easier because we might obtain a reasonably adequate baseline model performance by going through the models in our repository.

%% file: figures/schematic_diagram.tex
\begin{figure}[t]
  \centering
  \includegraphics[width=.95\linewidth]{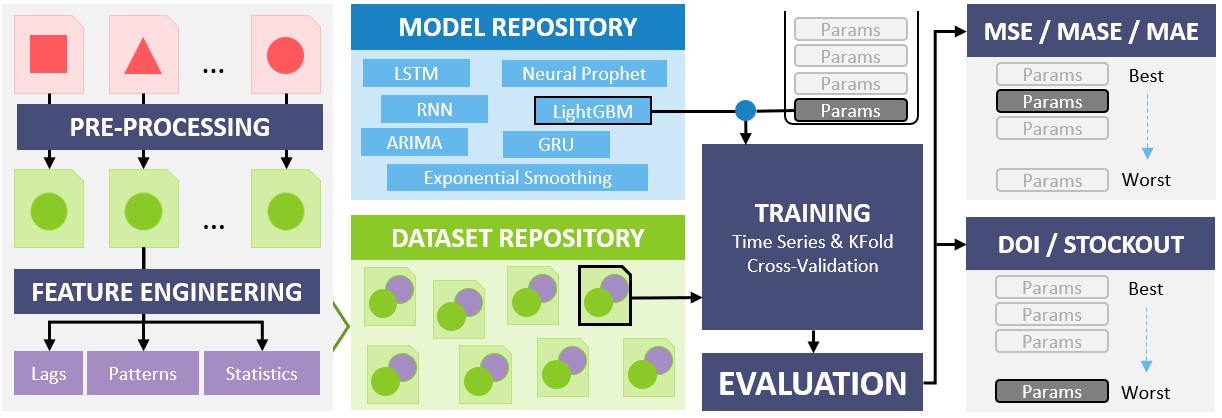}
  \vspace{-0.25cm}
  \caption{The proposed framework streamlines all development components of forecasting models.}
  \label{fig:schematic_diagram}
\end{figure}

%% file: figures/cross_validation_illustration.tex
\begin{figure}[t]
  \centering
  \includegraphics[width=.99\linewidth]{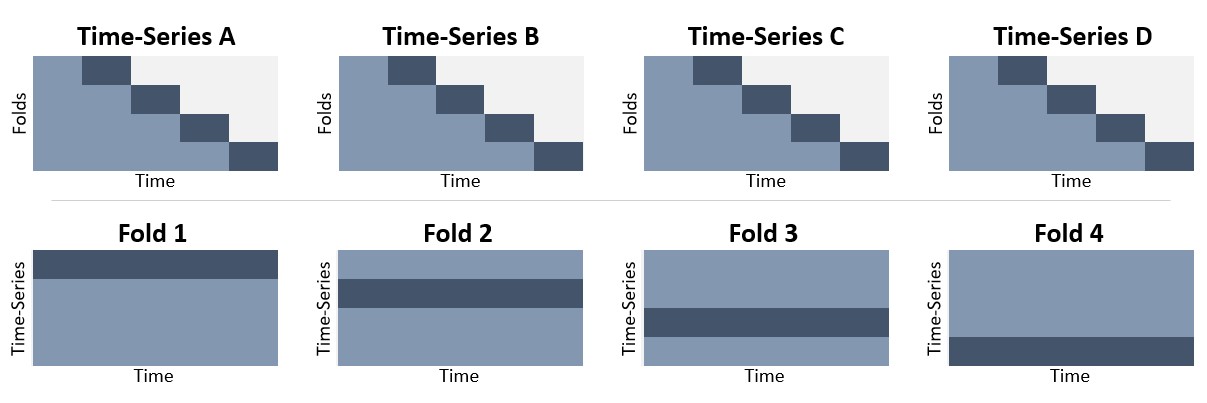}
    \caption{Illustration of time-series cross-validation (top) and k-fold cross-validation (bottom).}
  \label{fig:cross_validation_illustration}
\end{figure}

%% file: figures/doi_stockout_tradeoff.tex
\begin{figure}[t]
  \centering
    \includegraphics[width=.99\linewidth]{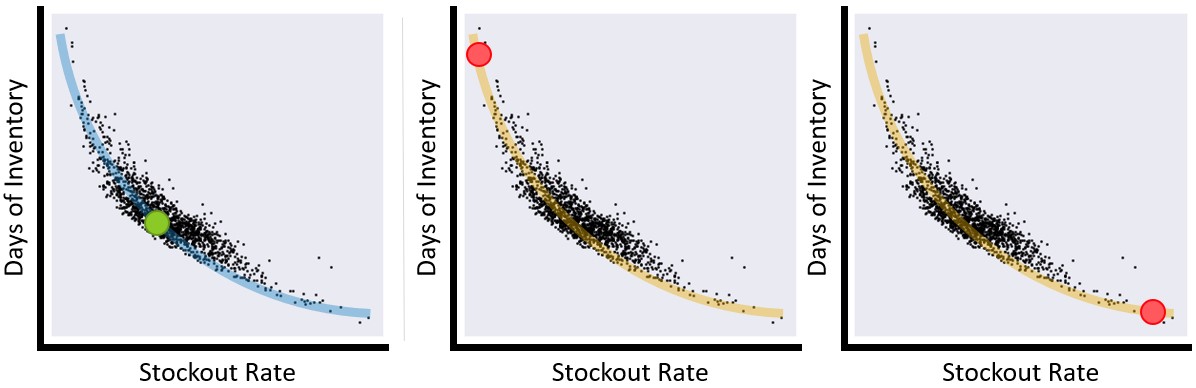}
    \caption{Illustration of optimization trade-off between DOI and SR (a) ideal; (b) low SR; (c) low DOI.}
    \vspace{0.5cm}
    \label{fig:doi_stockout_tradeoff}
\end{figure}

%% file: sections/04_case_study.tex
\section{Case study}
\label{sec:case-study}

In July 2020, we participated in an intelligent forecasting competition hosted by the USAID. The competition aimed to solve global health challenges by applying accurate forecasting algorithms for predicting the contraceptive demand and assisting with its distribution in Cote d'Ivoire. The competition was particularly challenging as it involved sizable time series data from 151 sites and 10 products, meaning that we were required to forecast the demand for more than one thousand unique time series. Given the monthly recording of logistics information, our goal was to predict the demand for the next three months.

To develop our models, we integrated the competition problem into our streamlined framework. By leveraging the capacities of our framework, we were able to quickly prototype and identify strong base models to work with. In Section \ref{sec:experiments}, we discuss the design choices for the proposed solution, beginning from training procedure, validation strategy, and model selection. We also included additional experiments to demonstrate the performance of such models when applied to an inventory management setting.



%% file: sections/05_experiments.tex
\section{Experiments}
\label{sec:experiments}
In this section, our goal is to evaluate the critical decisions that contributed to building our solution for the USAID competition rather than describing its exact implementation. For this purpose, we included three additional time-series datasets, each carrying a unique property of a real-world scenario. Using these datasets, we first assessed our decision to use a single model for all available time series (cross-learning) rather than developing a specific model for each of them. Next, we demonstrated a robust validation strategy for tuning hyperparameters of the models. Finally, we evaluated the best models selected via conventional metrics in an inventory management setting.



\input{tables/dataset_overview}
\subsection{Datasets}
\label{sec:exp_datasets}
To investigate the effectiveness of the design choices of our USAID competition solution, we considered three additional datasets, in addition to the USAID contraceptive dataset. These datasets are publicly available. The \textbf{USAID Contraceptive} dataset\footnote{\url{https://zindi.africa/competitions/usaids-intelligent-forecasting-challenge-model-future-contraceptive-use/data}} \citep{usaid-data} posed a challenging real-world problem because it contains various unique time series recorded at a monthly frequency. Note that the publicly available USAID contraceptive dataset is slightly different from the one used in the competition. The \textbf{Dairy Products} dataset\footnote{\url{https://mpr.datamart.ams.usda.gov/}} \citep{dairy-data} records the weekly consumption of five dairy products in the United States. This dataset represents a situation in which there are a few time series, each with a high temporal granularity. We also included the \textbf{Walmart Sales} dataset\footnote{\url{https://www.kaggle.com/c/m5-forecasting-accuracy/data}} of the M5 forecasting competition \citep{walmart-data,makridakis2021m5A}. We aggregated the recorded sales quantity by product group and site to achieve a dataset containing a sizable number of unique time series. Next, the \textbf{Kaggle Store Demand} dataset\footnote{\url{https://www.kaggle.com/c/demand-forecasting-kernels-only/data}} \citep{kaggle-data,bojer2021kaggle} consists of 500 products with weekly recording frequency, resulting in the most extensive dataset used in our experimental setting. We reserved three months of data points from each dataset as a separate-unseen test subset that is used to evaluate the models. Table \ref{tab:dataset_overview} provides a detailed description of each dataset.

\subsection{Single forecasting model using cross-learning}
\label{sec:exp_single_model}
The challenges of the USAID intelligent forecasting competition were related to the large number of time series. Although the logistics data were recorded across four years, from 2016 to 2020, they only have a monthly frequency. This implies that we were required to handle various unique time series, each of which is relatively short. This problem became more complicated with missing values and intermittent demand in large portions of the time-series data.

\input{tables/single_multi_model_comparison}

Statistical forecasting models are generally applied by relying on information from individual time series. Thus, we tuned multiple models equivalent to the number of series, further referred to as the multiple-model setup. Performing hyperparameter tuning on all the models would be infeasible because each set of hyperparameters needs to be applied across all models. An alternative solution was to consider AI-based forecasting models similar to the winning solution of the M5 forecasting competition. Hence, we accommodated several time series in a single-model setup, also referred to as cross-learning. This setup offers an immediate benefit where the AI model can learn patterns across different time series, which was highly suitable for the USAID contraceptive dataset. We leveraged the number of time series available in the dataset while removing the complexity of training one model for each. Therefore, iterating through a significantly larger number of hyperparameters became more straightforward.

We compared the two setups, single (cross-learning) and multiple models, using \textbf{NeuralProphet} \citep{taylor2018forecasting,triebe2020} and LightGBM \citep{ke2017lightgbm}. NeuralProphet only accepts a single time series dataset as the input. Therefore, it is limited to the multiple-model setup. The same inherent constraint would also apply to the statistical models. By contrast, we implemented both single and multiple model strategies for LightGBM. Here, \textbf{LightGBM-Single} represents the experimental setting in which we use the LightGBM with the single-model setup. The \textbf{LightGBM-Multi} further represents an experiment in which we assign one model for each time series.

As expected, the multiple-model setup yields more accurate predictions for datasets containing few, but long time-series data. The multiple-model setup introduces more model parameters to learn the pattern of each time series, which eventually translate to better accuracy. However, doing so might lead to severe overfitting problems when the number of samples is insufficient. This observation is highlighted by the lower errors in the dairy products dataset, but higher in the USAID dataset (see Table \ref{tab:single_multi_model_comparison}). As also observed by \citet{spiliotis2020comparison, bojer2021kaggle,makridakis2021m5B}, the single-model (i.e., cross-learning) setup enables the algorithm to extract and share underlying patterns across all available time series in the dataset. This ability helps the model tackle the main challenges of the USAID dataset because it compensates the low temporal granularity with the number of time series.


\input{tables/cross_validation}

\input{figures/lgbm_k_analysis}

\subsection{Multiple validation strategy}
\label{sec:exp_validation_strategy}

To evaluate the effectiveness of using the multiple cross-validation strategy, we sampled 3 K configurations for each LightGBM and LSTM, both with the single-model setup. For brevity, we have omitted the single-model suffix from the name of the models. Figure \ref{fig:lgbm_k_analysis} provides a quick overview of the prediction accuracy of the top-k configurations according to MASE because MASE is a scale independent metric that allows us to compare various datasets. As highlighted by the performance of LightGBM on the Kaggle dataset, by examining a smaller number of experiments, the top configurations exhibit a larger performance variance on the unseen test set. Therefore, it is essential to consider additional configurations when choosing the best model, as the top models are sensitive to small changes or even suffer from overfitting issues.

Next, we sliced Figure \ref{fig:lgbm_k_analysis} by observing the top-1000 configurations and presented the results in Table \ref{tab:cross_validation}. As shown in the table, using the time series cross-validation, we can identify configurations that transfer well between seen validation and the unseen test set. However, since the time series cross-validation only evaluates the configurations exclusively based on predicting future values, it fails to generalize across different time series. This problem is highlighted by the best performance on the validation set, which does not transfer to the test portion of the Walmart dataset.  

Our multiple cross-validation strategy incorporated both time series and k-fold cross-validation. Therefore, it can robustly identify configurations that can accurately predict future values while maintaining generalization for patterns across time series. As a result, we can consistently identify configurations that adapt well to the test set for all datasets, even when there exists a significant discrepancy between the validation and test, as highlighted by the Kaggle dataset (see Table \ref{tab:cross_validation}). This gap between the validation and test set is expected, as we only reserve a small number of data points within the unseen test set.




\subsection{Assessing forecasting models for inventory management setting}
\label{sec:exp_inventory}

Next, we investigated the performance of the models chosen using conventional regression metrics when applied to the inventory management setting. To this end, we applied the DOI and SR inventory metrics (defined in Equations \ref{eq:doi} and \ref{eq:sr}, respectively) in addition to the conventional metrics. As demonstrated in Table \ref{tab:regression_to_inventory}, selecting accurate models based on MASE led to a decent trade-off between DOI and SR. This property is expected as, by providing accurate predictions, the model performs relatively well on both DOI and SR. Similar behavior was found with the other conventional metrics (MAE and RMSE), which were omitted for the sake of brevity.

Although the conventional metrics allow us to choose accurate models, they do not indicate whether the errors stem from under, or overestimation of future demands. Moreover, they do not guarantee that the chosen model satisfies the desired inventory constraints. For example, in the global health setting represented by the USAID contraceptive dataset, we might favor models with the best SR to ensure that customers' demands were met. We can incorporate this user preference while choosing the best models by leveraging the flexibility of the framework in applying various evaluation metrics. In summary, training the models with the conventional metrics would enable them to provide an accurate estimate, while the DOI and SR would select models with predictions that favor the inventory setting, rather than pure accuracy.


\input{tables/regression_to_inventory}

%% file: tables/dataset_overview.tex
\begin{table}[t!]
    \centering
    \caption{Brief description of each cleaned dataset included in our experiments.}
    \begin{tabular}{|l|c|c|c|c|c|}
    \hline
    \textbf{Name}         & \textbf{Frequency}   & \textbf{Length}    & \textbf{\#Time-Series} \\ \hline
    USAID Contraceptive   & Monthly     & 42        & 762 \\
    Dairy Products        & Weekly      & 484       & 5   \\ 
    Walmart Sales         & Weekly      & 278       & 30  \\ 
    Kaggle Store Demand   & Weekly      & 261       & 500 \\ \hline
    \end{tabular}
    \label{tab:dataset_overview}
\end{table}

%% file: tables/single_multi_model_comparison.tex

\begin{table}[t]
    \caption{Comparison between NeuralProphet and LightGBM using single and multiple model strategy. }
    \centering
    \begin{tabular}{|l|c|c|c|c|c|}
    \hline
    \textbf{Metric}         & \textbf{Model}    & \textbf{USAID}     & \textbf{Dairy}     & \textbf{Walmart}   & \textbf{Kaggle} \\ \hline
    \multirow{3}{*}{MAE}    & NeuralProphet     & 14.5859 & 5935891.8020 & 809.0128 & 31.5787 \\
                            & LightGBM-Multi    & 13.6166 & \textbf{5559450.1860} & 734.5936 & 32.2843 \\
                            & LightGBM-Single   & \textbf{11.3646} & 5742281.9593 & \textbf{590.5159} & \textbf{30.3952} \\
    \hline
    \multirow{3}{*}{RMSE}   & NeuralProphet     & 31.7738 & 10784627.0688 & 1516.4113 & 46.6282 \\
                            & LightGBM-Multi    & 29.2691 & \textbf{9022726.5642} & 1472.2758 & \textbf{42.8675} \\
                            & LightGBM-Single   & \textbf{26.2613} & 9068631.7436 & \textbf{1032.0995} & 43.4232 \\
    \hline
    \multirow{3}{*}{MASE}   & NeuralProphet     & 1.5528 & 0.9199 & 1.2261 & 1.0803 \\
                            & LightGBM-Multi    & 1.4205 & \textbf{0.9145} & 1.0559 & 1.0863 \\
                            & LightGBM-Single   & \textbf{1.1616} & 0.9433 & \textbf{0.8969} & \textbf{1.0136} \\
    \hline
    \end{tabular}
    
    \label{tab:single_multi_model_comparison}
\end{table}

%% file: tables/cross_validation.tex

\begin{table}[ht!]
    \caption{Comparison between different cross-validation strategies on MAE, RMSE, and MASE.}
    \centering
    \setlength\tabcolsep{3pt} 
    \begin{adjustbox}{angle=270}
    \resizebox{!}{6.7cm}{
    \begin{tabular}{|c|c|c|cc|cc|cc|cc|}
    \hline
    \multirow{2}{*}{\textbf{Metric}} & \multirow{2}{*}{\textbf{Model}}& \multirow{2}{*}{\textbf{CV}} & \multicolumn{2}{c|}{\textbf{USAID}} & \multicolumn{2}{c|}{\textbf{Dairy}} & \multicolumn{2}{c|}{\textbf{Walmart}} & \multicolumn{2}{c|}{\textbf{Kaggle}} \\ 
          & & & Valid & Test & Valid & Test & Valid & Test & Valid & Test \\ \hline
          
    \multirow{6}{*}{MAE}  & \multirow{3}{*}{LGBM} & TS & 10.1634 & 11.8230 & 7227133.4748 & 5756410.0923 & \textbf{564.9514} & 584.4730 & 35.0562 & 22.5980 \\
                          &                       & KFold & 10.1651 & 11.8315 & 7237897.8031 & 5740567.3344 & 584.9778 & 583.0270 & 35.1731 & 22.4844 \\
                          &                       & Both & \textbf{10.1514} & \textbf{11.8093} & \textbf{7207972.2908} & \textbf{5685080.2655} & 566.4891 & \textbf{575.7160} & \textbf{35.0449} & \textbf{22.4148} \\
    \cline{2-11}
                          & \multirow{3}{*}{LSTM} & TS & 11.8949 & 13.0825 & 7552314.1023 & 6305839.8578 & 663.6378 & 710.5549 & 34.8898 & 28.9381 \\
                          &                       & KFold & 11.9651 & 12.9743 & 7609099.4352 & 6265020.7515 & 678.8025 & 709.8739 & 35.8380 & 27.7679 \\
                          &                       & Both & \textbf{11.8843} & \textbf{12.9477} & \textbf{7519657.1878} & \textbf{6214551.6917} & \textbf{656.8424} & \textbf{695.0386} & \textbf{34.4677} & \textbf{27.5839} \\
    \hline
    \multirow{6}{*}{RMSE} & \multirow{3}{*}{LGBM} & TS & 26.2004 & 26.0029 & 12255448.2921 & 8791346.7036 & 910.2738 & 1069.3699 & 46.1781 & 30.3938 \\
                          &                       & KFold & 26.2146 & 26.0558 & 12281805.1658 & 8785695.6910 & 934.3512 & 1084.9685 & 46.3201 & 30.2436 \\
                          &                       & Both & \textbf{26.1487} & \textbf{25.9906} & \textbf{12229970.6146} & \textbf{8759418.5526} & \textbf{909.6075} & \textbf{1066.7050} & \textbf{46.1634} & \textbf{30.1608} \\
    \cline{2-11}
                          & \multirow{3}{*}{LSTM} & TS & 27.7264 & 27.1020 & 12616295.4649 & 9822605.7652 & 1103.1000 & 1263.4808 & 46.6570 & 39.0600 \\
                          &                       & KFold & 27.8067 & 27.0024 & 12724024.0935 & 9749545.8742 & 1134.8565 & 1251.3391 & 47.9005 & 37.4096 \\
                          &                       & Both & \textbf{27.7019} & \textbf{26.9958} & \textbf{12568829.5047} & \textbf{9697273.7960} & \textbf{1088.5752} & \textbf{1228.7584} & \textbf{46.0689} & \textbf{37.0581} \\
    \hline
    \multirow{6}{*}{MASE} & \multirow{3}{*}{LGBM} & TS & 0.8698 & 1.1973 & 0.8654 & 1.0141 & \textbf{0.9251} & 0.8906 & 1.3546 & 0.7637 \\
                          &                       & KFold & 0.8921 & 1.2075 & 0.8663 & 1.0153 & 0.9820 & 0.8679 & 1.3590 & 0.7600 \\
                          &                       & Both & \textbf{0.8646} & \textbf{1.1861} & \textbf{0.8641} & \textbf{1.0020} & 0.9349 & \textbf{0.8575} & \textbf{1.3544} & \textbf{0.7578} \\
    \cline{2-11}
                          & \multirow{3}{*}{LSTM} & TS & 1.1932 & 1.3487 & 0.9066 & 1.1320 & 1.0602 & 1.1276 & 1.3471 & 0.9691 \\
                          &                       & KFold & 1.2170 & 1.3055 & 0.9125 & 1.1259 & 1.0848 & 1.1264 & 1.3845 & 0.9308 \\
                          &                       & Both & \textbf{1.1929} & \textbf{1.3013} & \textbf{0.9031} & \textbf{1.1114} & \textbf{1.0505} & \textbf{1.1032} & \textbf{1.3317} & \textbf{0.9242} \\
    \hline

    \end{tabular}
    }
    \end{adjustbox}
    \label{tab:cross_validation}
\end{table}

%% file: figures/lgbm_k_analysis.tex
\begin{figure}[t]
  \centering
  \includegraphics[width=.99\linewidth]{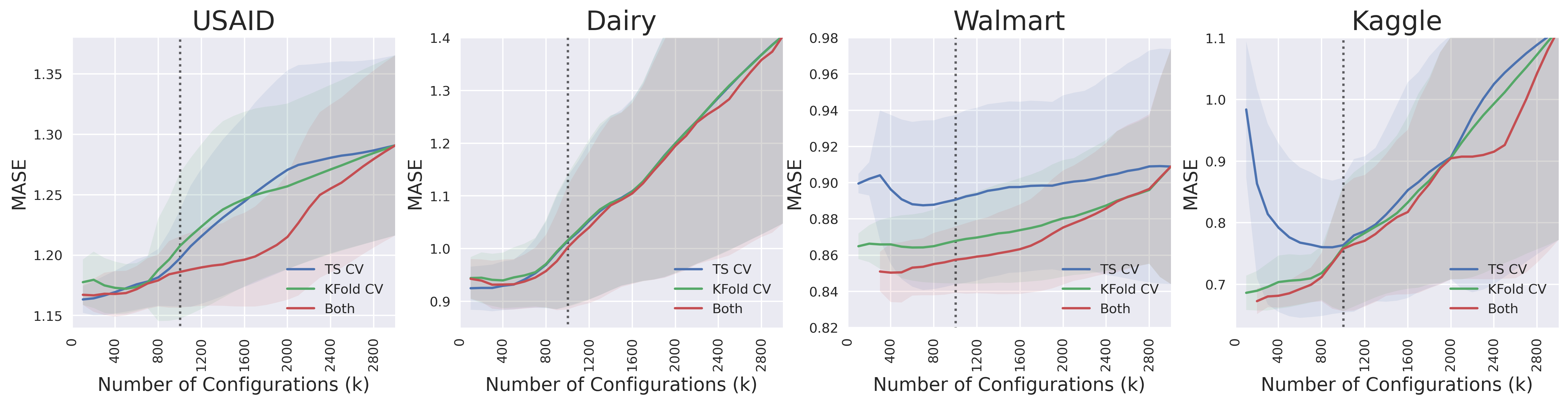}
\caption{Average MASE of cross-validation strategies computed from the top-k configurations of LightGBM.}
  \label{fig:lgbm_k_analysis}
\end{figure}

%% file: tables/regression_to_inventory.tex

\begin{table}[ht]
    \caption{Performance analysis of the best models based on MASE, DOI, and SR. }
    \centering
    \begin{tabular}{|l|l|ccc|ccc|}
    
    \hline 
    
    \multirow{2}{*}{\textbf{Dataset}} & \multirow{2}{*}{\textbf{Best of}} & \multicolumn{3}{c|}{\textbf{LightGBM}} & \multicolumn{3}{c|}{\textbf{LSTM}} \\
    & & \textbf{MASE} & \textbf{SR (\%)} & \textbf{DOI} & \textbf{MASE} & \textbf{SR (\%)} & \textbf{DOI} \\ 
    
    \hline \multirow{3}{*}{USAID}   & MASE        & \textbf{1.1365} & 45.8812 & 28.3914 & \textbf{1.2576} & 41.4141 & 39.2569 \\
                                    & SR (\%)     & 1.3384 & \textbf{34.1126} & 42.6878 & 1.3571 & \textbf{37.6723} & 44.7749 \\
                                    & DOI         & 1.3859 & 61.6276 & \textbf{20.0022} & 1.3994 & 48.9253 & \textbf{33.5962} \\
                                    
    \hline \multirow{3}{*}{Dairy}   & MASE        & \textbf{0.8482} & 3.9102 & 7.2427 & \textbf{0.9915} & 3.9031 & 7.3538 \\
                                    & SR (\%)     & 0.9686 & \textbf{3.3607} & 7.3982 & 1.2849 & \textbf{3.0090} & 7.7407 \\
                                    & DOI         & 0.9463 & 4.8160 & \textbf{7.1397} & 1.2195 & 5.9958 & \textbf{7.1614} \\
                                    
    \hline \multirow{3}{*}{Walmart} & MASE        & \textbf{0.8207} & 3.4908 & 6.8754 & \textbf{0.9750} & 3.5652 & 6.9265 \\
                                    & SR (\%)     & 0.8317 & \textbf{3.4210} & 6.8890 & 1.1509 & \textbf{2.6013} & 7.1289 \\
                                    & DOI         & 3.4659 & 21.4416 & \textbf{5.5041} & 2.7908 & 17.0041 & \textbf{5.8377} \\
                                    
    \hline \multirow{3}{*}{Kaggle}  & MASE        & \textbf{0.6522} & 3.2703 & 6.9292 & \textbf{0.7656} & 3.1037 & 7.0138 \\
                                    & SR (\%)     & 0.7620 & \textbf{2.3686} & 7.1183 & 1.4179 & \textbf{1.2799} & 7.6287 \\
                                    & DOI         & 1.5432 & 8.2296 & \textbf{6.7172} & 1.5374 & 10.2071 & \textbf{6.4397} \\
    \hline
    \end{tabular}
    
    \label{tab:regression_to_inventory}
\end{table}

%% file: sections/06_conclusion.tex
\section{Conclusion}
\label{sec:conclusion}

We proposed a streamlined framework to facilitate the development of forecasting models in an agile fashion by streamlining the core components of the process. The framework is equipped with the functionality of applying suitable pre-processing; therefore, it can resolve variations in format, issues, and features into a uniform interface. This uniform interface for all datasets is proven to be an essential aspect of the framework, as it allows users to extract better feature representations of time-series data and enables the same procedure to fit different algorithms. Users can further iterate through various training configurations, including feature representations, algorithms, and other hyperparameters. Furthermore, we included a novel combination of time series and k-fold cross-validation strategies into the framework to robustly identify configurations with the finest generalization properties. In addition, the framework integrates custom evaluation metrics to fit the task at hand. An example of this evaluation flexibility is the possibility of assessing the forecasting models directly in an inventory management setting through a simulation process. With these capabilities, the framework enables users to effortlessly incorporate new datasets, experiment on different algorithms, and select the best model configurations using the desired evaluation metrics.

Throughout the experiments, we discussed the design decisions and the application of our framework on different forecasting datasets. The USAID forecasting competition is our referent study case. We also demonstrated the advantages of cross-learning a single model from multiple time series. Additionally, the comparison of different validation strategies highlighted that our combination of multiple validations is beneficial for the selection of the best models. Finally, we evaluated the models with the highest accuracy using inventory metrics, thus providing insights into their performance in the inventory management setting, which allowed us to impose preferences on model selection.

The proposed framework prompts interesting research possibilities. For example, we could extend the application of 1-model and N-datasets to achieve better model generalization or even improve the model performance over shorter time series data. Another interesting direction would be to promote the sample diversity of an ensemble process considering subsets of the time-series data, similar to the idea of our adapted k-fold cross-validation.